\documentclass{article}

\usepackage{booktabs}
\usepackage{geometry}
\usepackage{multirow}
\usepackage[utf8]{inputenc}
\newcommand{\footremember}[2]{%
    \footnote{#2}
    \newcounter{#1}
    \setcounter{#1}{\value{footnote}}%
}
\newcommand{\footrecall}[1]{%
    \footnotemark[\value{#1}]%
} 

\title{Identifying Melanoma Images using EfficientNet Ensemble: Winning Solution to the SIIM-ISIC Melanoma Classification Challenge}
\author{%
  Qishen Ha \footremember{equal}{Equal contribution} \\ \small LINE Corp%
  \and 
  Bo Liu\footrecall{equal} \footnote{Corresponding author. Email: boli@nvidia.com} \\ \small NVIDIA%
  \and 
  Fuxu Liu\footrecall{equal} \\ \small ReadSense Ltd%
  }

\date{\today}

\usepackage{natbib}
\usepackage{graphicx}

\begin{document}
\maketitle

\begin{abstract}
We present our winning solution\footnote{The code is available at https://github.com/haqishen/SIIM-ISIC-Melanoma-Classification-1st-Place-Solution} to the SIIM-ISIC Melanoma Classification Challenge. It is an ensemble of convolutions neural network (CNN) models with different backbones and input sizes, most of which are image-only models while a few of them used image-level and patient-level metadata. The keys to our winning are: (1) stable validation scheme (2) good choice of model target (3) carefully tuned pipeline and (4) ensembling with very diverse models. The winning submission scored 0.9600 AUC on cross validation and 0.9490 AUC on private leaderboard. 
\end{abstract}

\section{Introduction}

Melanoma is the most deadly form of skin cancer, but it can be cured with minor surgery if caught early enough. Fast and accurate diagnosis could tremendously benefit doctors and patients. Recent advancements in deep learning based computer vision have pushed model performance to be close to (or exceed in some cases) human expert level in many medical fields. Besides accuracy, model based diagnosis has the advantage of easy scaling up and could be more accessible than doctors in certain regions. At a minimum, it could be a great assistance in a dermatologist's toolbox.

In SIIM-ISIC Melanoma Classification challenge, competitors are asked to build models to identify melanoma using images of skin lesions and metadata. More than 3300 teams participated in the competition. In this paper, we present our winning solution.

For a deep learning model to work well, it usually requires three things: (1) large amount of data, (2) big models, (3) competitive model design and training techniques. Our solution addresses point 1 by using both image data and metadata from both last year and this year's competitions. Point 2 is answered by a diverse group of state-of-the-art models (mostly EfficientNets \cite{efficientnet}) trained with various input sizes and setup. Finally, we solve point 3 by good choices of validation strategy and classification target.

In section 2, we explain the validation strategy in more details. Model architecture and the use of metadata are discussed in section 3. We then describe our augmentations and training setup in section 4. Lastly, the ensemble and scores are presented in section 5.

\section{Validation strategy}

In any machine learning project, it is critical to set up a trustworthy validation scheme, in order to properly evaluate and compare models. This is especially true if the dataset is small to medium size, or the evaluation metric is unstable, which is the case of this competition.

There are 33k images in train data. However, only 1.76\% are positive samples (i.e., malignant). The small number of positives causes the AUC metric to be very unstable, even with 5-fold cross validation.

Our solution to this problem is to use both last year (including 2018 and 2019) and this year's data (2020), and do 5-fold cross validation on the combined data. Even though last year's data is smaller (25k), it has 10 times (17.85\%) the positive sample ratio, making the AUC much more stable. We refer to this metric as $\textbf{cv\_all}$. For each experiment, we also track $\textbf{cv\_2020}$ as a sanity check. It is calculated on the 2020 data only (even though the model is trained on combined data).

The unstable score problem is even worse on public leaderboard (LB), with its data size being only one tenth of the training set. Throughout the competition, we did not use public LB score feedback in any way when developing models. In fact, we found that the public LB scores are so noisy that its correlation with $\textbf{cv\_all}$ is essentially 0 for the single models we had submitted. For ensembles, this problem is alleviated to a certain degree, as the noise from single models are averaged out. The bigger the ensemble, the more stable the LB score.

\section{Model architecture and metadata}

For a typical image classification problem, the standard approach is to take a deep CNN model (such as the most popular EfficientNet) trained on ImageNet, replace the last layer so that the output dimension equals the target's dimension, and fine tune it on the specific dataset.

Our winning solution follows the same road map with two twists: using diagnosis as target and adding metadata in some models.

The target to predict in this year's competition is binary---benign (i.e. no melanoma) or malignant (i.e. melanoma). We noticed that the target information is included in the diagnosis column: target is malignant if and only if diagnosis is melanoma. But diagnosis column is more granular when an image is benign. We believed using diagnosis as target to train the model could give the model more information.

The fact that diagnosis was the target to predict in last year's competition makes this choice more logical. There is a small problem though. The set of diagnosis is different between this year and last year. We solved it by mapping this year's diagnosis to last year's according to the descriptions on last year's website\footnote{https://challenge2019.isic-archive.com/}. See Table \ref{tab:targets} for the mapping. There are 9 target values in most of our models. In one model, we only used 4 target values (NV, MEL, BKL and Unknown) by mapping the other five (*) to Unknown.\\

\begin{table*}[ht]
\begin{center}
\begin{tabular}{lll}
\toprule
2019 Diagnosis       & 2020 Diagnosis                     & Target                   \\ 
\toprule
NV                   & nevus                              & NV                       \\ \hline
MEL                  & melanoma                           & MEL                      \\ \hline
BCC                  &                                    & BCC*                      \\ \hline
\multirow{4}{*}{BKL} & seborrheic keratosis               & \multirow{4}{*}{BKL}     \\
                     & lichenoid keratosis                &                          \\
                     & solar lentigo                      &                          \\
                     & lentigo NOS                        &                          \\ \hline
AK                   &                                    & AK*                       \\ \hline
SCC                  &                                    & SCC*                      \\ \hline
VASC                 &                                    & VASC*                     \\ \hline
DF                   &                                    & DF*                       \\ \hline
                     & cafe-au-lait macule                & \multirow{2}{*}{Unknown} \\
                     & atypical melanocytic proliferation &                          \\ 
\bottomrule                     
\end{tabular}
\caption{\textbf{Mapping from diagnosis to targets.} }\label{tab:targets}
\end{center}
\end{table*}

This means that the last layer of our classification model has 9-dimensional output. It is trained with cross entropy loss. When calculating AUC score and making submission, we take the MEL class's probability.

The second twist in the CNN model is the addition of 14 metadata features in some models: $\textbf{sex}$, $\textbf{age\_approx}$, 10 one-hot encoded $\textbf{anatom\_site\_general\_challenge}$ features, $\textbf{image\_size}$ in bytes and $\textbf{n\_images}$, where $\textbf{n\_images}$ is the number of all images of that patient in the data.

\begin{figure}[]
\centering
\includegraphics[width=1\textwidth]{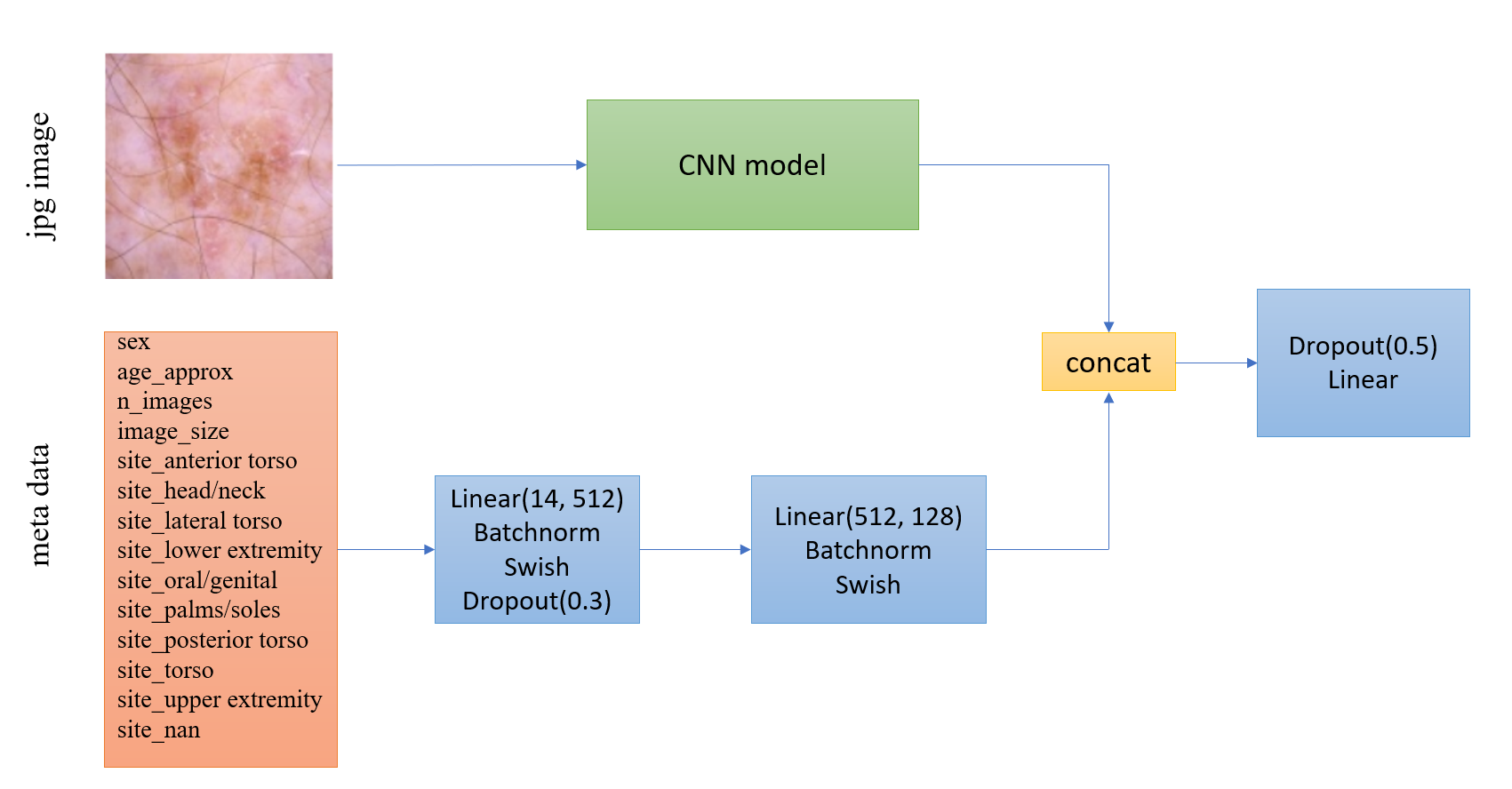}
\caption{\textbf{Model architecture of metadata models.} }
\label{fig:model}
\end{figure}

The metadata go through two fully connected layers before being concatenated with the CNN features, which then go to the final fully connected layer. The model architecture with metadata is illustrated in Figure \ref{fig:model}. In three of four metadata models, the two hidden layers have dimensions 512 and 128 as in the figure. In another metadata model, they are 128 and 32. We observe that models with images only perform better than metadata models overall, but the addition of of metadata models in the ensemble provides good model diversity.

\section{Augmentations and training setup}

\begin{figure}[]
\centering
\includegraphics[width=1\textwidth]{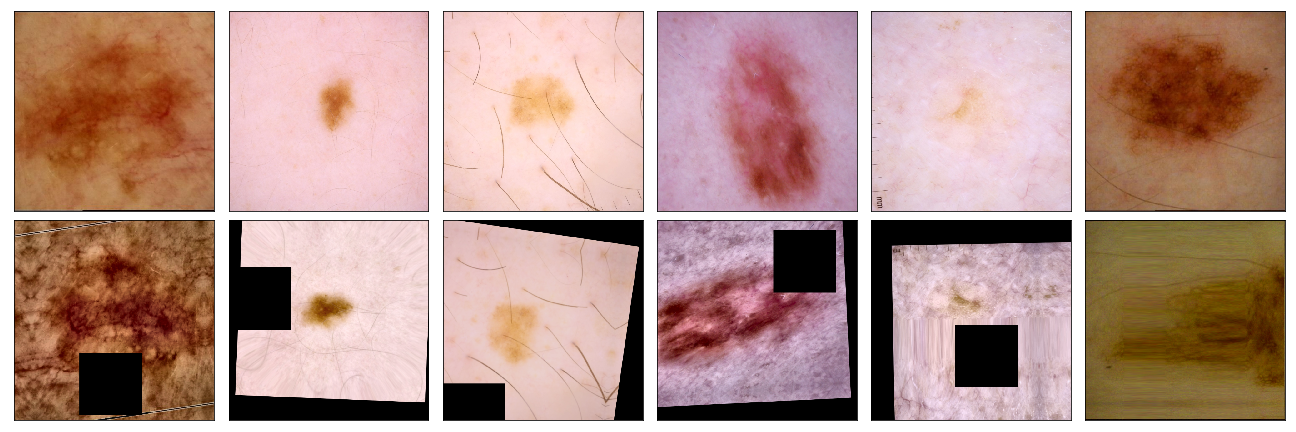}
\caption{Training augmentation of 6 random images. Top row: original images; bottom row: augmented images.}
\label{fig:aug}
\end{figure}

In small to medium sized datasets, image augmentation is important to prevent overfit. We used in our pipeline the following augmentations from the popular and powerful Pytorch augmentation library Albumentations \cite{albu} : Transpose, Flip, Rotate, RandomBrightness, RandomContrast, MotionBlur, MedianBlur, GaussianBlur,
GaussNoise, OpticalDistortion, GridDistortion, ElasticTransform, CLAHE, HueSaturationValue,
ShiftScaleRotate, Cutout\cite{cutout}. Figure \ref{fig:aug} is an illustration of the before-after of these augmentations.

For training schedule, we used cosine annealing with one warm up epoch \cite{cosine}. The total number of epochs is 15 for most models. The initial learning rate of the cosine cycle is tuned for each model, which ranges from $1e-4$ to $3e-4$. The learning rate in the warm up epoch is always one tenth of the initial learning rate of the cosine cycle. Batch size is 64 for all models. All training were done on NVIDIA Tesla V100 GPUs in mixed precision. Up to 8 GPUs were used in parallel.

\section{Ensemble}

In order to make the ensemble stronger, we trained a variety of diverse models. 18 models were selected to comprise the final ensemble as listed in Table \ref{tab:score}. The model diversity comes from multiple angles.
\begin{itemize}
    \item Backbones: EfficientNet B3, B4, B5, B6, B7, SE-ResNeXt-101 \cite{senet}, ResNeSt-101 \cite{resnest}
    \item Different targets: Model 9 has only 4 classes
    \item 5 folds are split differently among models
    \item Resized image input sizes: 384, 448, 512, 576, 640, 768 and 896
    \item 4 of the 18 models used metadata with different hidden layer dimensions
\end{itemize}

The scores of the 18 single models demonstrate that the stability of the 4 metrics are: $\textbf{cv\_all} > \textbf{cv\_2020} > \textbf{private\_LB} > \textbf{public\_LB}$, as measured by their respective standard deviations 0.0012, 0.0043, 0.0060, 0.0093. 

The final ensemble is a simple average of the 18 models' probability ranks. In other words, we transform each model's probability predictions to uniform [0,1] before averaging them.

\begin{table*}[ht]
\begin{center}
\small
\setlength\tabcolsep{1.5pt} 
\begin{tabular}{clcccccccccc}
\toprule
\multicolumn{1}{l}{Model}    & Backbone & Target & \multicolumn{1}{l}{Input} & \multicolumn{1}{l}{Resize} & Metadata & \multicolumn{1}{l}{Init lr} & \multicolumn{1}{l}{Epochs} & \multicolumn{1}{l}{cv\_all} & \multicolumn{1}{l}{cv\_2020} & \multicolumn{1}{l}{private\_LB} & \multicolumn{1}{l}{public\_LB} \\
\hline
1                            & B3       & 9c     & 768                       & 512                        & yes      & 3e-5                       & 18                         & 0.9762                         & 0.9300                      & 0.9305                         & 0.9182                        \\
2                            & B4     & 9c     & 768                       & 640                        &          & 3e-5                       & 15                         & 0.9767                         & 0.9400                      & 0.9299                         & 0.9342                        \\
3                            & B4     & 9c     & 768                       & 768                        &          & 3e-5                       & 15                         & 0.9771                         & 0.9408                      & 0.9264                         & 0.9251                        \\
4                            & B4     & 9c     & 768                       & 640                        & yes      & 3e-5                       & 15                         & 0.9765                         & 0.9408                      & 0.9302                         & 0.9221                        \\
5                            & B4     & 9c     & 1024                      & 896                        &          & 2e-5                       & 15                         & 0.9744                         & 0.9390                      & 0.9320                          & 0.9281                        \\
6                            & B4     & 9c     & 512                       & 448                        &          & 3e-5                       & 15                         & 0.9748                         & 0.9307                      & 0.9213                         & 0.9002                        \\
7                            & B5     & 9c     & 512                       & 384                        & yes      & 3e-5                       & 15                         & 0.9752                         & 0.9329                      & 0.9167                         & 0.9350                        \\
8                            & B5     & 9c     & 768                       & 640                        &          & 1.5e-5                     & 15                         & 0.9771                         & 0.9428                      & 0.9291                         & 0.9216                        \\
9                            & B5     & 4c     & 768                       & 640                        &          & 1.5e-5                     & 15                         & 0.9765                         & 0.9384                      & 0.9362                         & 0.9260                        \\
10                           & B5     & 9c     & 512                       & 448                        &          & 3e-5                       & 15                         & 0.9751                         & 0.9397                      & 0.9363                         & 0.9387                        \\
11                           & B6     & 9c     & 768                       & 640                        &          & 3e-5                       & 15                         & 0.9756                         & 0.9444                      & 0.9408                         & 0.9283                        \\
12                           & B6     & 9c     & 768                       & 576                        &          & 3e-5                       & 15                         & 0.9761                         & 0.9443                      & 0.9266                         & 0.9245                        \\
13                           & B6     & 9c     & 512                       & 448                        &          & 3e-5                       & 15                         & 0.9742                         & 0.9383                      & 0.9261                         & 0.9154                        \\
14                           & B7     & 9c     & 512                       & 384                        & yes      & 3e-5                       & 15                         & 0.9748                         & 0.9394                      & 0.9193                         & 0.9170                        \\
15                           & B7     & 9c     & 768                       & 576                        &          & 1e-5                       & 15                         & 0.9764                         & 0.9432                      & 0.9260                          & 0.9271                        \\
16                           & B7     & 9c     & 768                       & 640                        &          & 1e-5                       & 15                         & 0.9754                         & 0.9440                      & 0.9304                         & 0.9115                        \\
17                           & SE\_X101 & 9c     & 768                       & 640                        &          & 3e-5                       & 15                         & 0.9739                         & 0.9428                      & 0.9295                         & 0.9337                        \\
18                           & Nest101 & 9c     & 768                       & 640                        &          & 2e-5                       & 15                         & 0.9728                         & 0.9396                      & 0.9320                          & 0.9267                        \\
\hline
\multicolumn{1}{l}{Ensemble} &          &        & \multicolumn{1}{l}{}      & \multicolumn{1}{l}{}       &          & \multicolumn{1}{l}{}        & \multicolumn{1}{l}{}       & \textbf{0.9845 }                        & \textbf{0.9600}                      & \textbf{0.9490}                         & \textbf{0.9442}    \\                   
\bottomrule
\end{tabular}
\caption{\textbf{Model configuration and scores.} Model 1 to 16 are EfficientNets. Model 17 is SE-ResNext101. Model 18 is ResNest101. All models have 9 classes except model 9 which has 4 classes. Images of dimension ``Input'' are read from the disk then resized into ``Resize'' dimension before being fed to the model. ``Init lr'' is the learning rate of warm up epoch. All models are trained for 15 epochs except for Model 1 which is trained for 18.}\label{tab:score}
\end{center}
\end{table*}

\bibliographystyle{plain}
\bibliography{references}
\end{document}